\documentclass{article}
\usepackage{palatino}
\usepackage[margin=1in]{geometry}
\usepackage{listings}
\usepackage{lipsum}
\usepackage{natbib}
\usepackage{courier}
\usepackage[utf8]{inputenc}
\usepackage{float}
\usepackage[T1]{fontenc}    
\usepackage{hyperref}       
\usepackage{url}            
\usepackage{booktabs}       
\usepackage{amsfonts}       
\usepackage{nicefrac}       
\usepackage{microtype}      
\usepackage{graphicx}
\usepackage{wrapfig}
\usepackage{subcaption}
\usepackage{placeins}
\usepackage{hyperref}
\usepackage[english]{babel}
\usepackage{url}
\usepackage{mathtools}
\usepackage{amsmath}
\usepackage{amsthm}
\usepackage{amssymb}
\usepackage{wrapfig}
\usepackage{bm}
\usepackage{psfrag}
\usepackage{tikz}
\usepackage{epstopdf}
\usepackage{xcolor}
\usepackage{adjustbox}

\newcommand{\beq}[1][\vspace{0.3em}]{#1\begin{equation}}
\newcommand{\eeq}{\end{equation}}

\newcommand{\bit}{\vspace{0mm}\begin{itemize}}
\newcommand{\eit}{\vspace{0mm}\end{itemize}}
\newcommand{\ben}{\vspace{0mm}\begin{enumerate}}
\newcommand{\een}{\vspace{0mm}\end{enumerate}}

\newcommand{\bb}[1]{\mathbb{#1}}

\newcommand{\mc}[1]{\mathcal{#1}}

\definecolor{light-gray}{gray}{0.95}

\title{An Empirical Analysis of Proximal Policy Optimization with Kronecker-factored Natural Gradients}
\author{Jiaming Song \\ Stanford University \\ \href{tsong@cs.stanford.edu}{tsong@cs.stanford.edu} 
\and Yuhuai Wu \\ University of Toronto \\ \href{ywu@cs.toronto.edu}{ywu@cs.toronto.edu} }
\date{}

\begin{document}

\maketitle
\section{Introduction}
Deep reinforcement learning methods have shown tremendous success in a large variety tasks, such as Go~\citep{silver2016mastering}, Atari~\citep{mnih2013playing}, and continuous control~\citep{lillicrap2015continuous,schulman2015trust}. Policy gradient methods~\citep{williams1992simple} is an important family of methods in model-free reinforcement learning, and the current state-of-the-art policy gradient methods are Proximal Policy Optimization (~\citet{schulman2017proximal}) and ACKTR~\citep{wu2017scalable}. The two methods, however, take different approaches to better sample efficiency: PPO considers a particular ``clipping'' objective that mimics a trust-region, whereas ACKTR considers approximated natural gradients that balances speed and optimization.

In this technical report, we consider an approach that combines the PPO objective and K-FAC natural gradient optimization, for which we call PPOKFAC. We perform a range of empirical analysis on various aspects of the algorithm, such as sample complexity, training speed, and sensitivity to batch size and training epochs. We observe that PPOKFAC is able to outperform PPO in terms of sample complexity and speed in a range of MuJoCo environments, while being scalable in terms of batch size. In spite of this, it seems that adding more epochs is not necessarily helpful for sample efficiency, and PPOKFAC seems to be worse than its A2C counterpart, ACKTR.

\section{Background and Related Works}
\subsection{Reinforcement Learning}
We consider an agent agent interacting with a discounted Markov Decision Process with infinite horizon $(\mc{S}, \mc{A}, \gamma, P, r)$. The agent $\pi_\theta(a \vert s)$ selects an action $a_t \in \mc{A}$ from its policy given the state $s_t \in \mc{S}$ at time $t$. The environment produces a reward $r(s_t, a_t)$ and transitions to the next state $s_{t+1}$ according to the transition probability $P(s_{t+1} \vert s_t, a_t)$. The agent aims to maximize its $\gamma$-discounted cumulative reward 
$$
\mc{L}(\theta) = \bb{E}_{\pi}[\sum_{j=0}^{\infty} \gamma^i r(s_{t+i}, a_{t+i})]
$$
with respect to the policy parameters $\theta$, where $\bb{E}_\pi[\cdot]$ is the expectation over the trajectories created through executing the policies in the environment. 

\subsection{Policy Gradient Methods}
Policy gradient methods~\citep{williams1992simple} directly updates the parameters $\theta$ so as to maximize the objective $\mc{L}(\theta)$. In particular, the policy gradient is defined as:
$$
\nabla_\theta \mc{L}(\theta) = \bb{E}_\pi \big[\sum_{t=0}^{\infty} \nabla_\theta \pi_\theta(a_t \vert s_t) A(t) \big]
$$
where $A(t)$ is often chosen to be the advantage function $A^\pi(s_t, a_t)$. For example, the advantage actor critic methods defines the advantage function as $k$-step returns with function approximation,
$$
A^\pi(s_t, a_t) = \sum_{i=0}^{k-1} \big(\gamma^i r(s_{t+i}, a_{t+i}) + \gamma^k V_\phi^\pi(s_{t+k}) \big) - V_\phi^\pi(s_t)
$$
where $V_\phi^\pi$ is the value network that provides an estimate of the $\gamma$-discounted rewards from the policy $\pi$. 

\subsection{Trust Region Methods and Proximal Policy Optimization}
Trust regions methods propose to maximize the following ``surrogate'' objective:
$$
\bb{E}_{\pi}\Big[\frac{\pi_\theta(a_t \vert s_t)}{\pi_{\theta_{\text{old}}}(a_t \vert s_t)}A(t) \Big]
$$
while satisfying the following trust region condition: 
$$\bb{E}_\pi[D_{\text{KL}}(\pi_{\theta_{\text{old}}}(a_t \vert s_t), \pi_\theta(\cdot \vert s_t))] \leq \delta$$ where $\theta_{\text{old}}$ is the vector of policy parameters before the update, and $D_{\text{KL}}$ is the KL-divergence. In particular, the Proximal Policy Optimization (PPO, ~\citet{schulman2017proximal}) algorithm propose a modification to the objective which penalizes the new policy to be far from the old policy without explicitly enforcing the trust region constraint. The PPO objective is the following:
$$
\mc{L}^{\text{PPO}}(\theta) = \bb{E}_{\pi}\Big[\min(\frac{\pi_\theta(a_t \vert s_t)}{\pi_{\theta_{\text{old}}}(a_t \vert s_t)} A(t), \text{clip}(\frac{\pi_\theta(a_t \vert s_t)}{\pi_{\theta_{\text{old}}}(a_t \vert s_t)}, 1-\epsilon, 1+\epsilon) A(t)) \Big]
$$
where $\epsilon$ is a hyperparameter. The goal for the clipping term is to penalize policies updates that are too large, similar to the trust region constraint. This objective is then optimized through stochastic gradients descent (PPO-SGD).

\subsection{Scalable Natural Gradients using Kronecker-factored Approximation}
Natural gradients~\citep{amari1998natural} performs steepest descent over the metric constructed by the Fisher information matrix $F$, which is a local quadratic approximation of the KL divergence. Unlike the usual Euclidean norm used in stochastic gradient descent, this norm is independent of the model parameterization $\theta$ on the class of probability distributions. 

A recently proposed technique based on Kronecker-factored approximate curvature (K-FAC, \citet{martens2015optimizing}) uses a Kronecker-factored approximation to the Fisher matrix to perform efficient approximate natural gradient updates. K-FAC approxmates the Fisher matrix by first assuming independence between parameters at different layers and then approximating the local Fisher matrix through Kronecker factorization. This technique gives rise to a more efficient actor-critic algorithm called the Actor Critic using Kronecker-Factored Trust Region (ACKTR, ~\citet{wu2017scalable}), which has shown to significantly outperform its stochastic gradient descent counterpart (A2C, ~\citet{mnih2016asynchronous}).

\section{Proximal Policy Optimization with K-FAC Natural Gradients}
Both PPO and ACKTR are state-of-the-art in their respective regimes -- PPO obtains the advantage through the clipping objective function that can be easily optimized, while ACKTR achieves higher performance through (an approximation of) natural gradient updates. The significant contributing factor to their success, while difrerent in nature, are not mutually exclusive. Therefore, a naturally interesting question would be: 
\begin{quote}
How would an algorithm perform if we combine the advantages of both fronts?
\end{quote}
In this technical report, we attempt to investigate this question through some empirical analysis.

\subsection{Natural Gradient in Proximal Policy Optimization}
Similar to the ACKTR approach, we consider the Fisher information matrix over the policy function, which defines a distribution over the actions given the current state and take the expectation over the trajectory function:
$$
F = \bb{E}_{\tau}\Big[\nabla_\theta \log \pi(a_t \vert s_t) (\nabla_\theta \log \pi(a_t \vert s_t))^\top\Big]
$$
where $\bb{E}_{\tau}$ is the expectation over trajectories.
For the critic, we consider the Gauss-Newton matrix, which is equivalent to the Fisher for a Gaussian observation model~\footnote{See more details in ~\citet{wu2017scalable}.}.

\subsection{Adaptive Selection of Learning Rate}
In the SGD case for PPO, the learning rate $\eta$ is selected according to a fixed linear decaying schedule. However, in the case of K-FAC (and second order optimization methods in general), it is often difficult to select a reasonable learning rate for large updates. Therefore, we adopt the trust region formulation of K-FAC and choose the learning rate according to the following schedule.

Suppose the trust region threshold is $\delta$:
\begin{itemize}
    \item If actual KL $\geq 2 \delta$, then $\eta := \eta / 1.5$.
    \item If actual KL $\leq \delta / 2$, then $\eta := 1.5 \eta$.
    \item Otherwise $\eta$ stays the same.
\end{itemize}
This schedule decreases learning rate when the new policy seems to be deviating far from the old one, while increases learning rate with the new policy is too close.

\subsection{Batch-size and Update Iterations}
In each outer loop of the PPOSGD algorithm, a large batch of experience is collected while the stochastic gradient optimizer operates on a smaller minibatch for a number of epochs. For example, in the Mujoco experiments in \cite{}, each batch consists of 2048 state-action pairs and is updated for 10 epochs with a SGD minibatch of size 64. In the ACKTR case, however, the K-FAC optimizer takes in a single large batch and updates only once. 

In the case of PPO with K-FAC, we consider the batch size schedule of ACKTR: feeding the optimizers large batches at a time with fewer number of updates. We find that this has superior performance empirically than updating with minibatches. Having larger batches seems to reduce the variance of the Fisher matrix estimation for K-FAC. Moreover, we consider updating the batch for $n$ epochs, which is basically $n$ updates. For example, if we consider $n=2$ and batch size 2048 (for the Mujoco environments), K-FAC would perform 2 updates, while SGD would perform $2048 / 64 \times 10 = 320$ updates. This suggests that one K-FAC update is much more efficient than one SGD update.

\section{Experiments}
We name the resulting algorithm PPOKFAC, and conduct a series of experiments on the MuJoCo environments, which aims to investigate the following questions:
\begin{itemize}
    \item How does PPOKFAC compare with PPOSGD in terms of sample complexity?
    \item How does the performance of PPOKFAC change in terms of batch size?
    \item How does the performance of PPOKFAC change in terms of number of epochs?
    \item How does PPOKFAC compare with PPOSGD in terms of speed?
\end{itemize}

For PPOSGD, we use the same hyperparameters in \cite{}, except for Humanoid-v1 environment where we set the learning rate to $10^{-4}$, which seems to give better performance. For PPOKFAC, $\eta = 0.03$ is the initial learning rate, $\delta=0.002$ is the trust region radius and batch size is $2048$, and the number of epochs is $2$ (unless specified otherwise). We use two layer fully connected neural networks with (64, 64) neurons for both actor and critic for all the tasks, except for Humanoid which we use (256, 256).

\subsection{Comparison with PPOSGD}
\begin{figure}
    \centering
    \includegraphics[width=\textwidth]{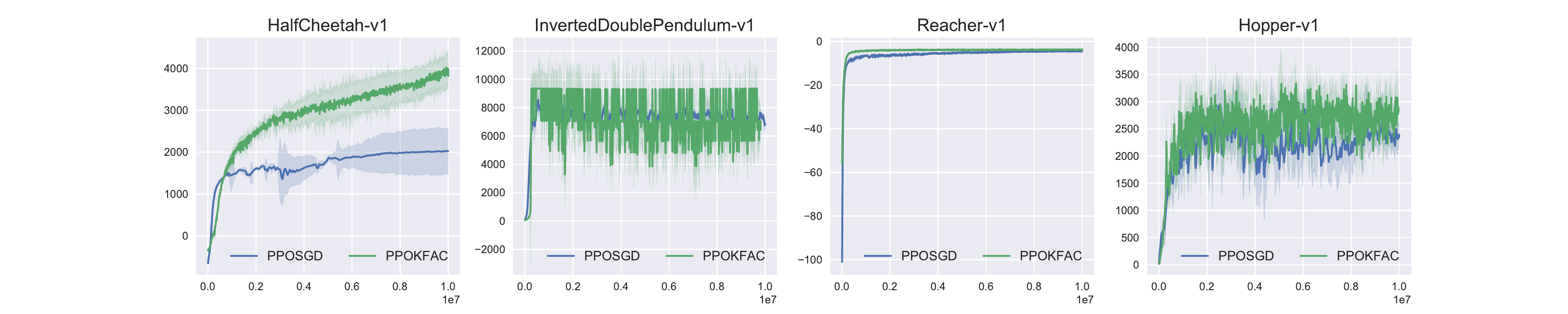}
    \caption{Sample complexity of PPOSGD and PPOKFAC on MuJoCo 10M benchmarks.}
    \label{fig:regular-sgd-compare}
\end{figure}

In Figure \ref{fig:regular-sgd-compare} we present the mean of rewards of the last 100 episodes in training as a function of training timesteps. Notably, PPOKFAC outperforms PPOSGD in HalfCheetah, Reacher and Hopper while having similar sample complexity in InvertedDoublePendulum.

\begin{figure}
    \centering
    \includegraphics[width=0.3\textwidth]{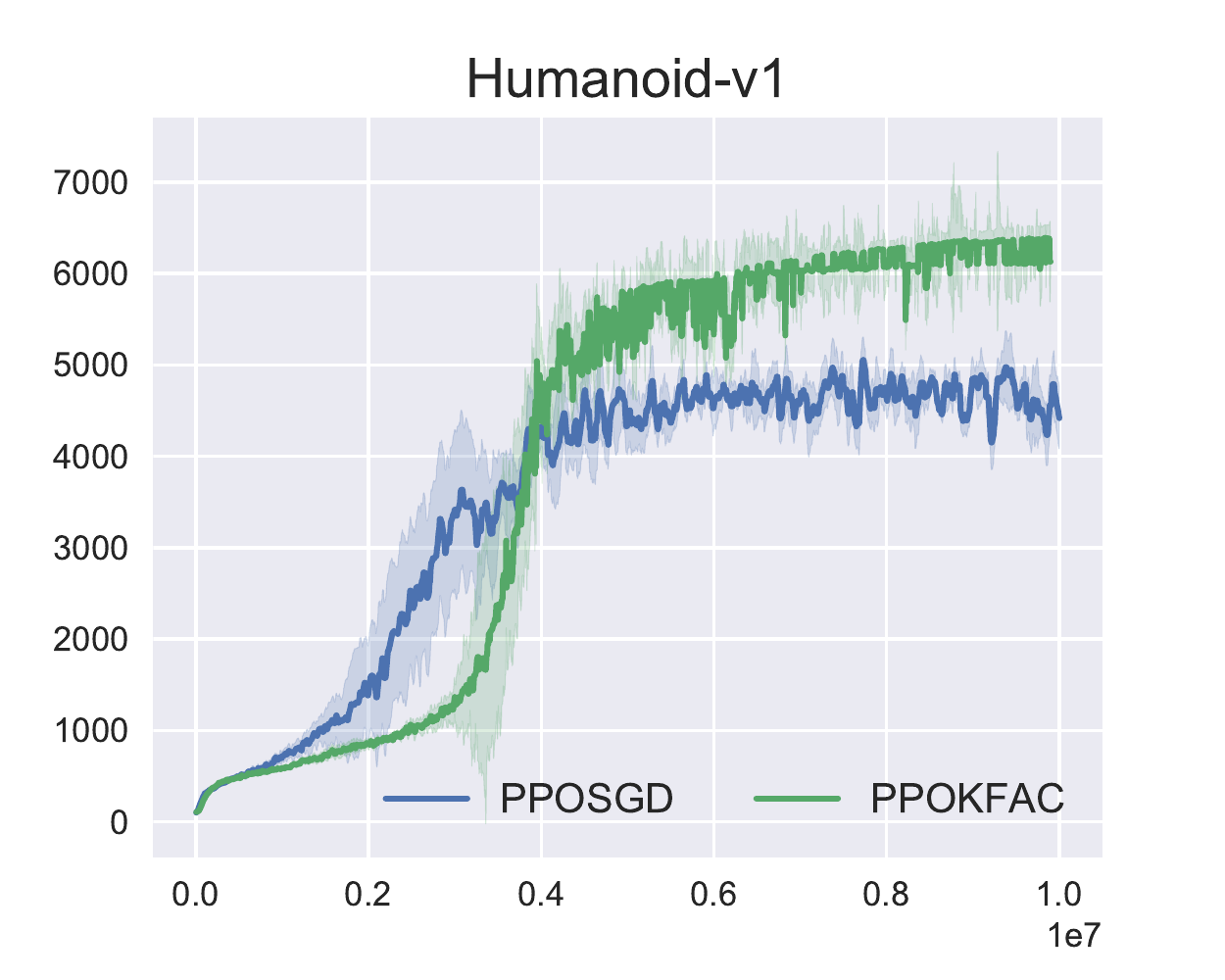}
    \caption{Sample complexity of PPOSGD and PPOKFAC on Humanoid.}
    \label{fig:humanoid-sgd-compare}
\end{figure}

Interestingly, in the HalfCheetah environment we observe that PPOKFAC underperforms at first but catches up and surpasses PPOSGD after a while. We also observe such phenomenon in Humanoid (in Figure \ref{fig:humanoid-sgd-compare}), where PPOKFAC surpasses PPOSGD only at around 4 million timesteps. 

\begin{figure}
    \centering
    \begin{subfigure}{0.3\textwidth}
        \includegraphics[width=\textwidth]{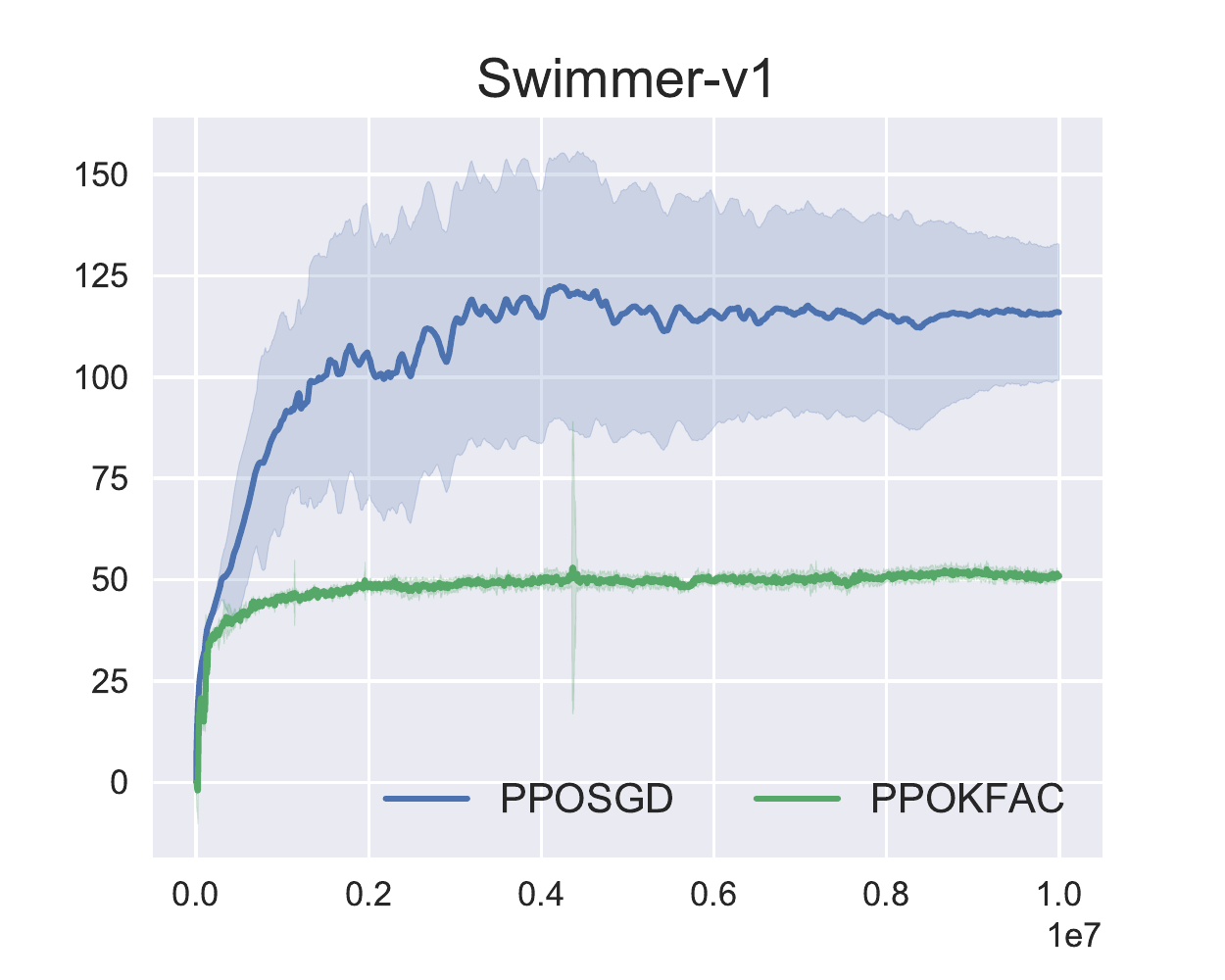}
        \caption{Learning rate is managed by trust region.}
    \end{subfigure}
    ~
    \begin{subfigure}{0.3\textwidth}
        \includegraphics[width=\textwidth]{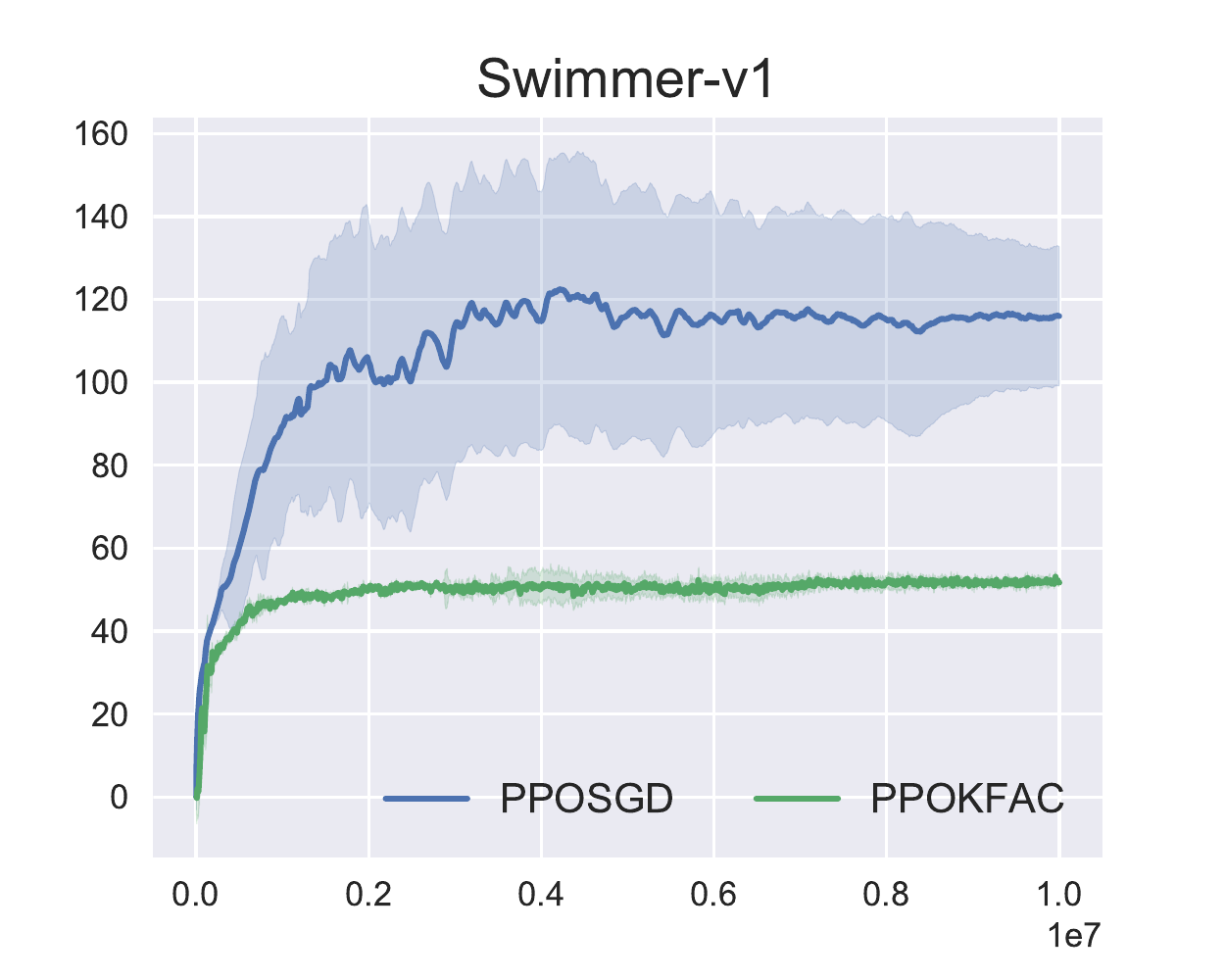}
        \caption{Learing rate is decreased linearly.}
    \end{subfigure}
    \caption{Performance comparsions between PPOSGD and PPOKFAC on Swimmer.}
    \label{fig:swimmer}
\end{figure}

In the Swimmer environment, however, we observe a different result (in Figure~\ref{fig:swimmer}) where PPOSGD vastly outperforms PPOKFAC. We also considered using a linearly decreasing learning rate schedule and also achieved similar results with PPOKFAC.

\subsection{Batch Size}
\begin{figure}
    \centering
    \begin{subfigure}{\textwidth}
        \includegraphics[width=\textwidth]{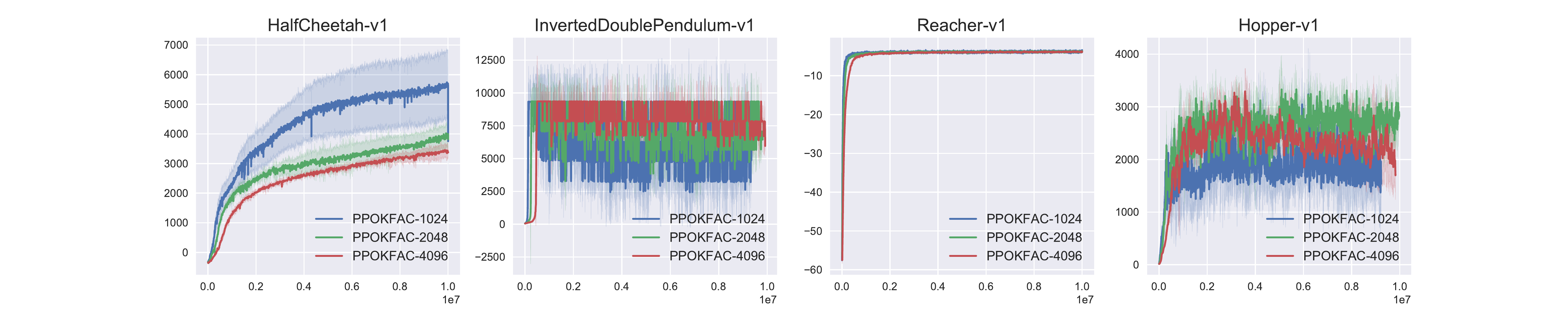}
        \caption{Timesteps as $x$ axis.}
    \end{subfigure}
    ~
    \begin{subfigure}{\textwidth}
        \includegraphics[width=\textwidth]{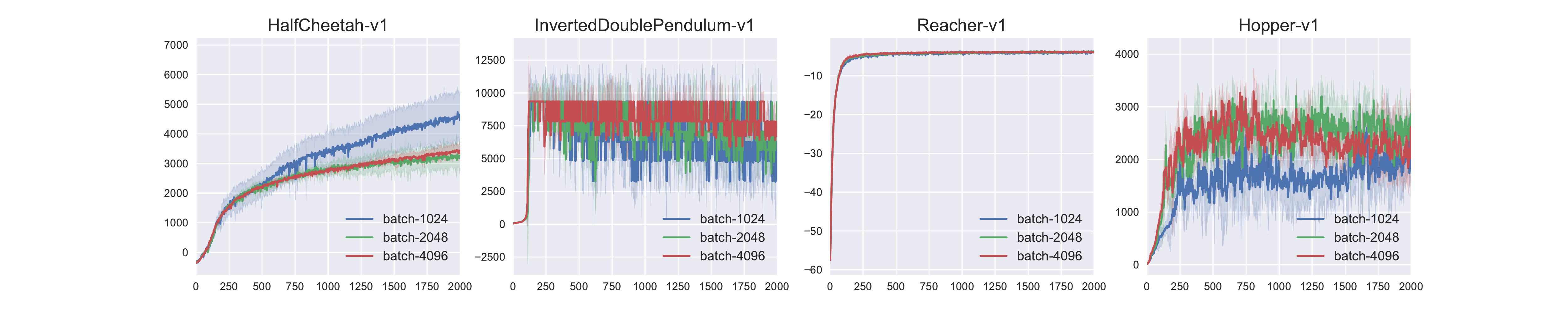}
        \caption{Updates as $x$ axis.}
    \end{subfigure}
    \caption{Performance comparsions between different batch sizes (regular MuJoCo benchmarks).}
    \label{fig:regular-batch}
\end{figure}

\begin{figure}
    \centering
    \begin{subfigure}{0.3\textwidth}
        \includegraphics[width=\textwidth]{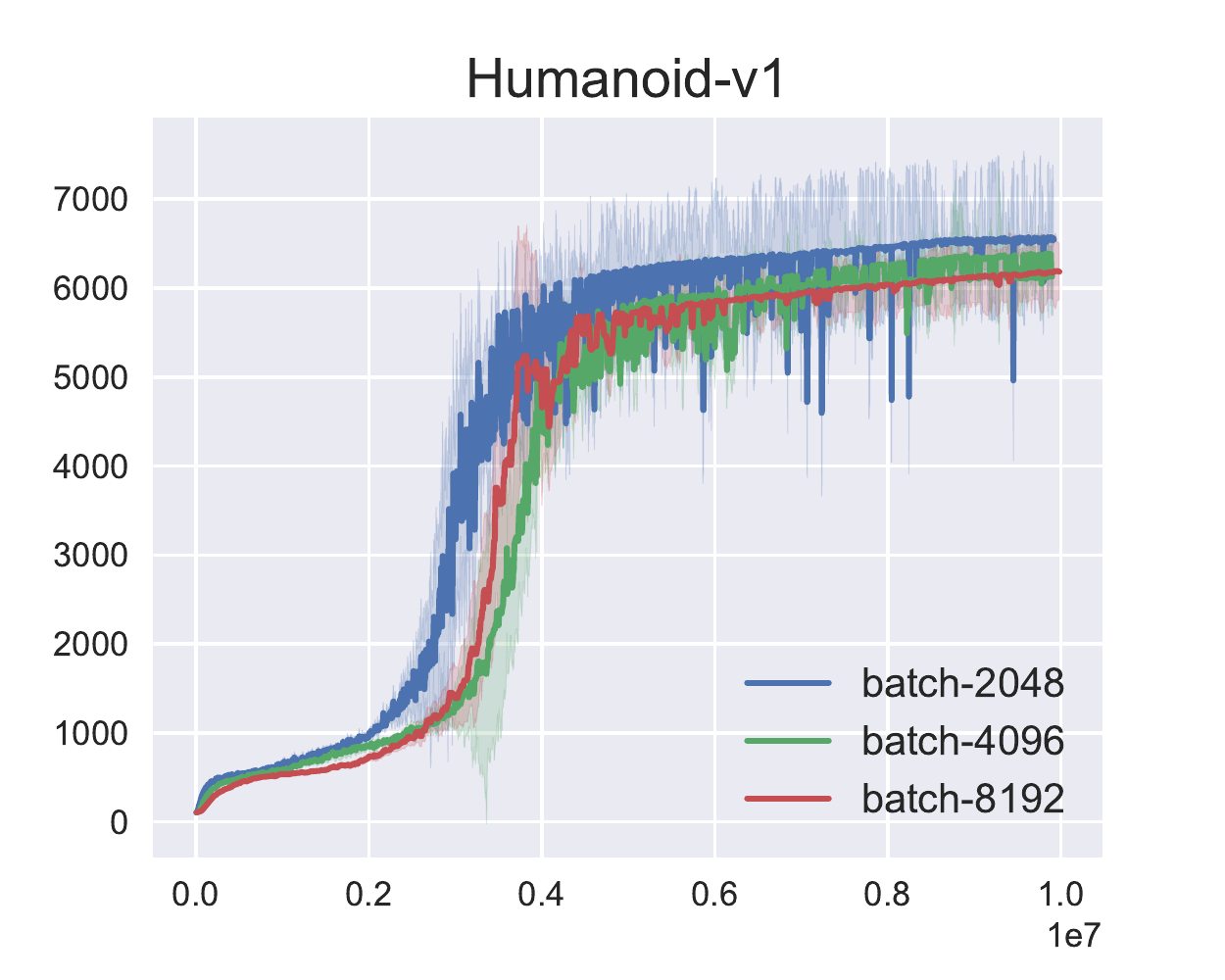}
        \caption{Timesteps as $x$ axis.}
    \end{subfigure}
    ~
    \begin{subfigure}{0.3\textwidth}
        \includegraphics[width=\textwidth]{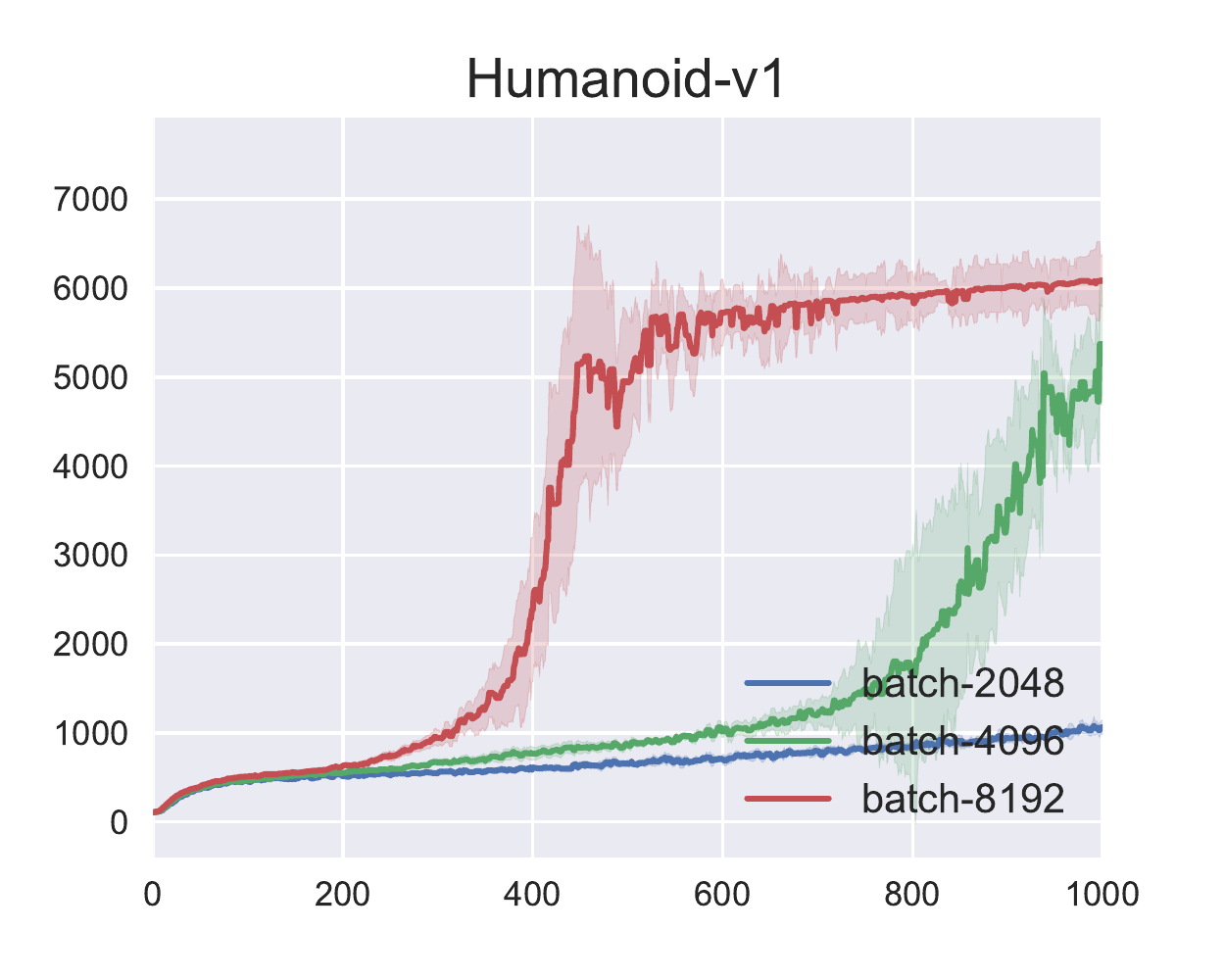}
        \caption{Updates as $x$ axis.}
    \end{subfigure}
    \caption{Performance comparsions over different batch sizes (Humanoid).}
    \label{fig:humanoid-batch}
\end{figure}

In this section, we compare how PPOKFAC would perform under different batch sizes. Figures \ref{fig:regular-batch} and \ref{fig:humanoid-batch} show the performance of MuJoCo environments under different batch sizes when compared with the number of timesteps and number of updates respectively.
We find that ACKTR performance is relatively stable under different batch sizes even at the same number of timesteps,
which suggests that updates with larger batches are more efficient.

It seems that a smaller batch size seems to provide better performance in terms of the number of timesteps, yet in a distributed settings the number of updates is often the bottleneck since larger batch sizes can be obtained by multiple machines in parallel.
In our experiments, the algorithm with batch size of 1024 would have two times the number of updates than one with a batch size of 2048, under the same number of timesteps, which gives small batches an advantage when comparing under the number of timesteps.

Interesingly, the efficiency of large batch sizes is most clearly demonstrated in the Humanoid environment, which is the most complex benchmark in MuJoCo. It is possible that the effect of large batch sizes is most significant when the batch size is large.

\subsection{Update Epochs}
It is unclear as to how many epochs to update with the K-FAC optimizer. If we only update PPOKFAC once, it is essentially the ACKTR algorithm: notice that in the first update, the ratio between $\pi$ and $\pi_{\text{old}}$ is $1$ and there is no clipping at the first update, and only after the first update would the clipping function have effect. 

\begin{figure}
    \centering
    \includegraphics[width=\textwidth]{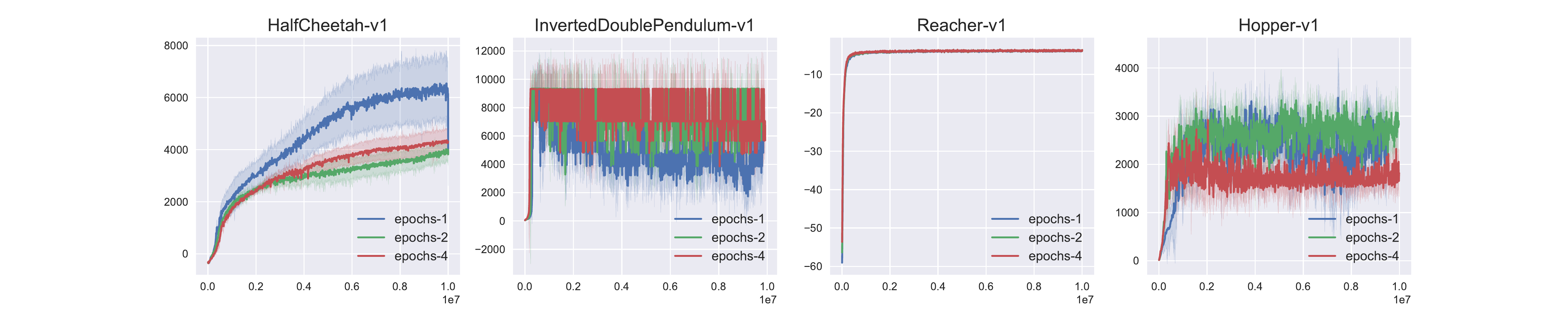}
    \caption{Performance comparisons over different epochs on MuJoCo 10M benchmarks.}
    \label{fig:regular-epoch}
\end{figure}

\begin{figure}
    \centering
    \includegraphics[width=0.3\textwidth]{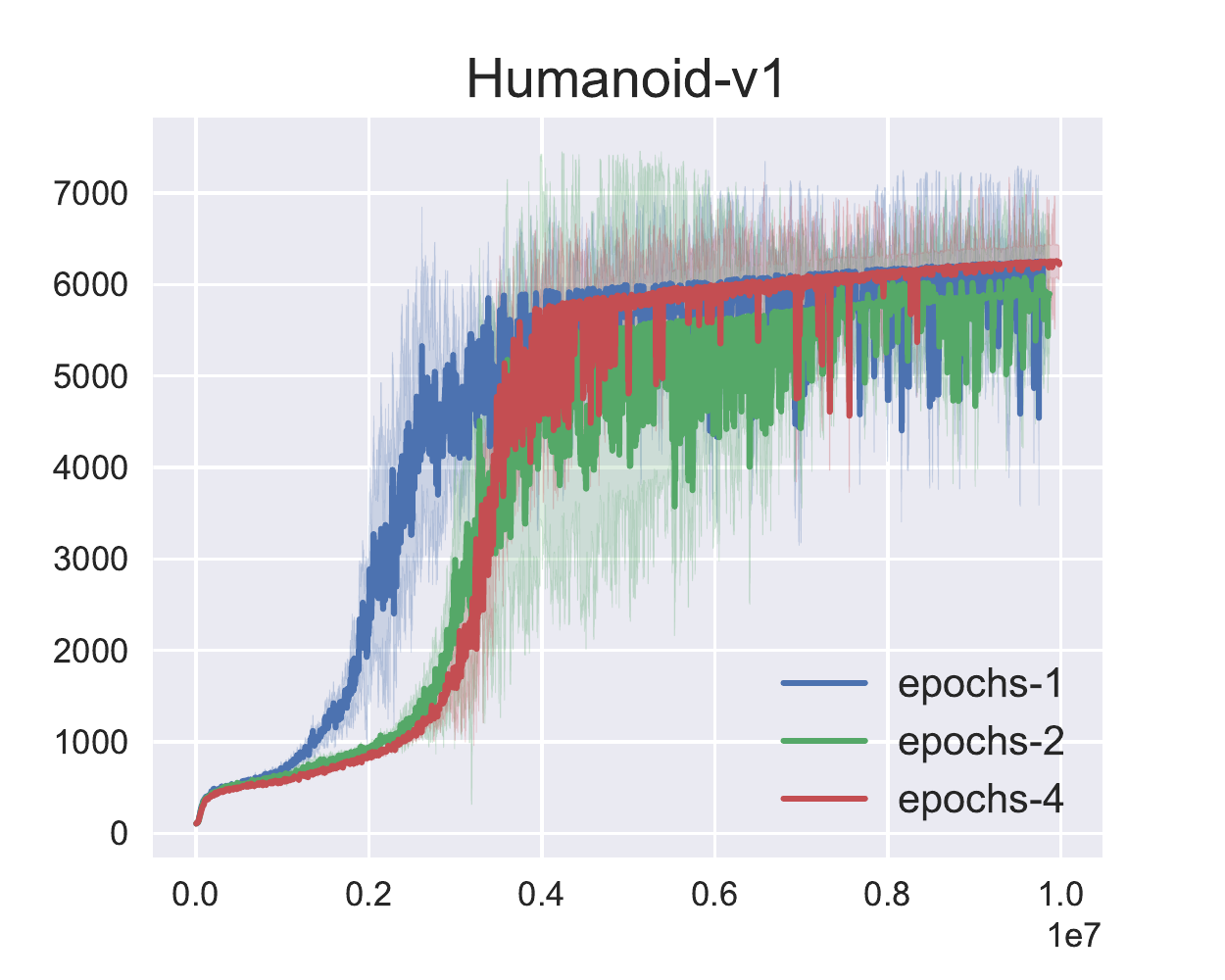}
    \caption{Performance comparisons over different epochs on Humanoid.}
    \label{fig:humanoid-epoch}
\end{figure}

Figures \ref{fig:regular-epoch} and \ref{fig:humanoid-epoch} show the results. Surprisingly, it seems that having more updates in PPOKFAC does not necessarily help sample complexity; in fact updating once (ACKTR) generally has the best performance. This might be caused by the fact that ACKTR has a component that enforces trust region which overlaps the effect from the PPO objective. Moreover, the large steps taken by K-FAC and the estimation of Fisher information matrix could also negatively affect performance by over-optimizing through multiple steps.

\subsection{Optimization Time}
One of the important factors of an algorithm is its computational complexity. Although in the case of many RL benchmarks, simulation is often the bottleneck rather than optimization, this will not necessarily be the case in a distributed setting, since simulations are often embarrisingly parallel. 

\begin{figure}
    \centering
    \includegraphics[width=0.3\textwidth]{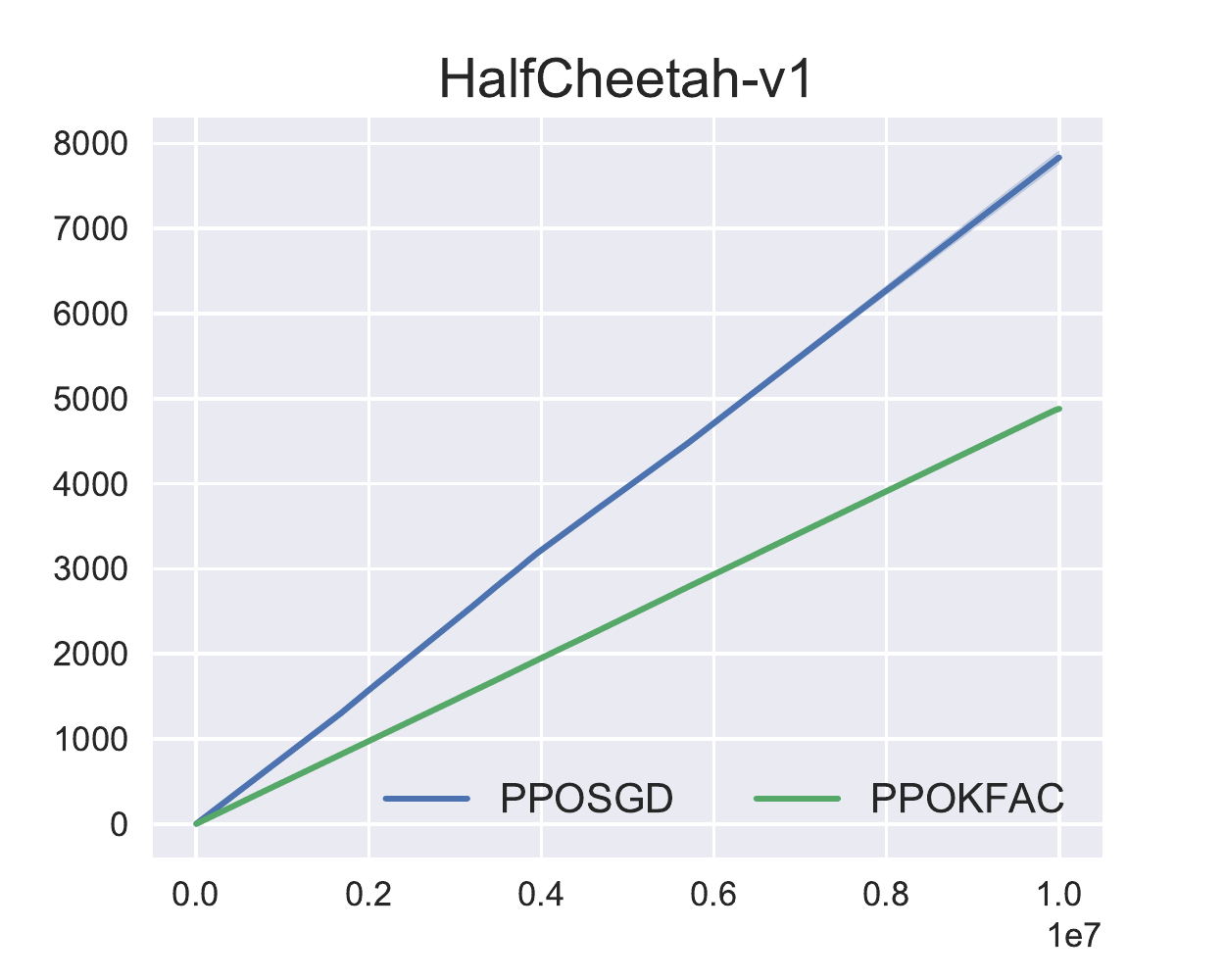}
    \caption{Time spent on optimization of PPOSGD and PPOKFAC.}
    \label{fig:halfcheetah-speed}
\end{figure}

Although being a (approximately) second order gradient update, K-FAC optimizers are competitive with first order optimizers (such as AdaM~\cite{kingma2014adam}) in terms of speed. This is further enhanced by the fact that K-FAC optimizers would generally require less updates than SGD optimizers do (2 epochs as opposed to 10). In Figure~\ref{fig:halfcheetah-speed}, we show the optimization time spent on a single CPU of the two algorithms. It turns out that the K-FAC approach is faster since it requires less epochs~\footnote{PPOSGD performance is much worse if trained with smaller number of epochs (such as 2).}. 

\section{Discussion}
While both PPO and ACKTR are state-of-the-art methods for deep RL, it seems that combining the advantages of both sides does not seem to improve sample complexity as expected; the objective and optimization procedure that works for SGD does not necessarily work for K-FAC. It would be interesting to explore the explanation behind this phenomenon. We assume that one reason is that the PPO objective implicitly defines a trust region through ratio clipping, and this overlaps with the learning rate selection criteria of ACKTR. Moreover, ACKTR takes much larger step-sizes than PPO and taking more iterations with such a step size might hurt performance (for example, in value function estimation).

\bibliographystyle{plainnat}
\bibliography{bib}
\end{document}